\documentclass{article}

\usepackage[final]{corl_2026}
\usepackage{xspace}
\usepackage{booktabs}
\newcommand{\alg}{\textcolor{black}{\textsc{TacEx}}\xspace}

\newcommand{\algdsrlsac}{\textsc{DSRL-SAC}\xspace}
\newcommand{\algdsrltacsac}{\textsc{DSRL-Tac-SAC}\xspace}
\newcommand{\algdsrlmaxinfosac}{\textsc{DSRL-MaxinfoSac}\xspace}
\newcommand{\algdsrltacmaxinfosac}{\textsc{DSRL-TacEx}\xspace}
\title{Tactile Curiosity Drives Robot Interaction}

\author{
  Klemens Iten\thanks{Equal contribution. Project website: \url{https://go.klem.nz/tacex}}\\
  ETH Z\"urich\\
  Switzerland\\
  \And
  Alexander Proshkin\footnotemark[1]\\
  University of California, Berkeley\\
  United States\\
  \And
  Bhavya Sukhija\footnotemark[1]\\
  ETH Z\"urich\\
  Switzerland\\
  \And
  Stelian Coros\\
  ETH Z\"urich\\
  Switzerland\\
  \And
  Andreas Krause\\
  ETH Z\"urich\\
  Switzerland\\
  \And
  Pieter Abbeel\\
  University of California, Berkeley\\
  United States\\
  \And
  Carmelo Sferrazza\\
  University of California, Berkeley\\
  United States\\
}

\usepackage{amsmath,amsfonts,bm}
\usepackage{xspace}
\usepackage{enumerate}
\usepackage{enumitem}
\usepackage{xfrac}
\usepackage{wrapfig}
\usepackage{parskip}
\usepackage{subcaption}
\usepackage{algorithm,algpseudocode}
\usepackage{pifont}
\usepackage[capitalize,noabbrev]{cleveref}

\algrenewcommand{\algorithmiccomment}[1]{\hfill\ding{228} #1}
\usepackage[para,online,flushleft]{threeparttable}
\usepackage{adjustbox}

\def\eqref#1{Equation~\ref{#1}}

\def\1{\bm{1}}

\def\va{{\bm{a}}}

\def\vk{{\bm{k}}}

\def\vs{{\bm{s}}}

\def\vv{{\bm{v}}}
\def\vw{{\bm{w}}}

\def\vz{{\bm{z}}}

\def\vpi{{\bm{\pi}}}
\def\vsigma{{\bm{\sigma}}}

\DeclareMathAlphabet{\mathsfit}{\encodingdefault}{\sfdefault}{m}{sl}
\SetMathAlphabet{\mathsfit}{bold}{\encodingdefault}{\sfdefault}{bx}{n}

\def\gA{{\mathcal{A}}}

\def\gS{{\mathcal{S}}}

\def\gW{{\mathcal{W}}}

\newcommand{\viota}{\boldsymbol{\iota}}   

\newcommand{\E}{\mathbb{E}}

\newcommand{\R}{\mathbb{R}}

\NewDocumentCommand{\norm}{sm}{\IfBooleanTF{#1}{\|#2\|}{\left\| #2 \right\|}}

\begin{document}
\maketitle


\begin{abstract}
\looseness=-1
Mastering robot manipulation skills via reinforcement learning (RL) remains largely sample-inefficient. The most common RL algorithms rely on random action sampling to discover new strategies, resulting in agents that allocate most of their training budget to motions in free space, away from the contacts from which manipulation skills emerge. Existing intrinsic motivation methods based on model disagreement or epistemic uncertainty improve on isotropic noise, but they can also reward uncertainty in functionally irrelevant transitions, such as erratic motions in free space. In this work, we argue that tactile feedback provides a natural signal for exploration, and introduce \alg, a framework that incorporates touch into epistemic uncertainty-driven exploration by decomposing model uncertainty across sensory modalities and directing curiosity toward the tactile channel. By anchoring curiosity to the sense of touch, \alg drives the robot to discover complex contact dynamics, learning to manipulate and grasp objects without task rewards or expert demonstrations during exploration. 
The interaction-dense dataset collected through this tactile-driven curiosity supports offline learning of downstream pick-and-place policies without additional environment interaction. We further use tactile-driven exploration to post-train vision-language-action (VLA) models. Although the VLAs are initially pre-trained without tactile feedback, post-training with \alg substantially improves downstream performance while remaining highly sample-efficient.
\end{abstract}



\section{Introduction}

\begin{figure}
    \centering
    \includegraphics[width=\linewidth]{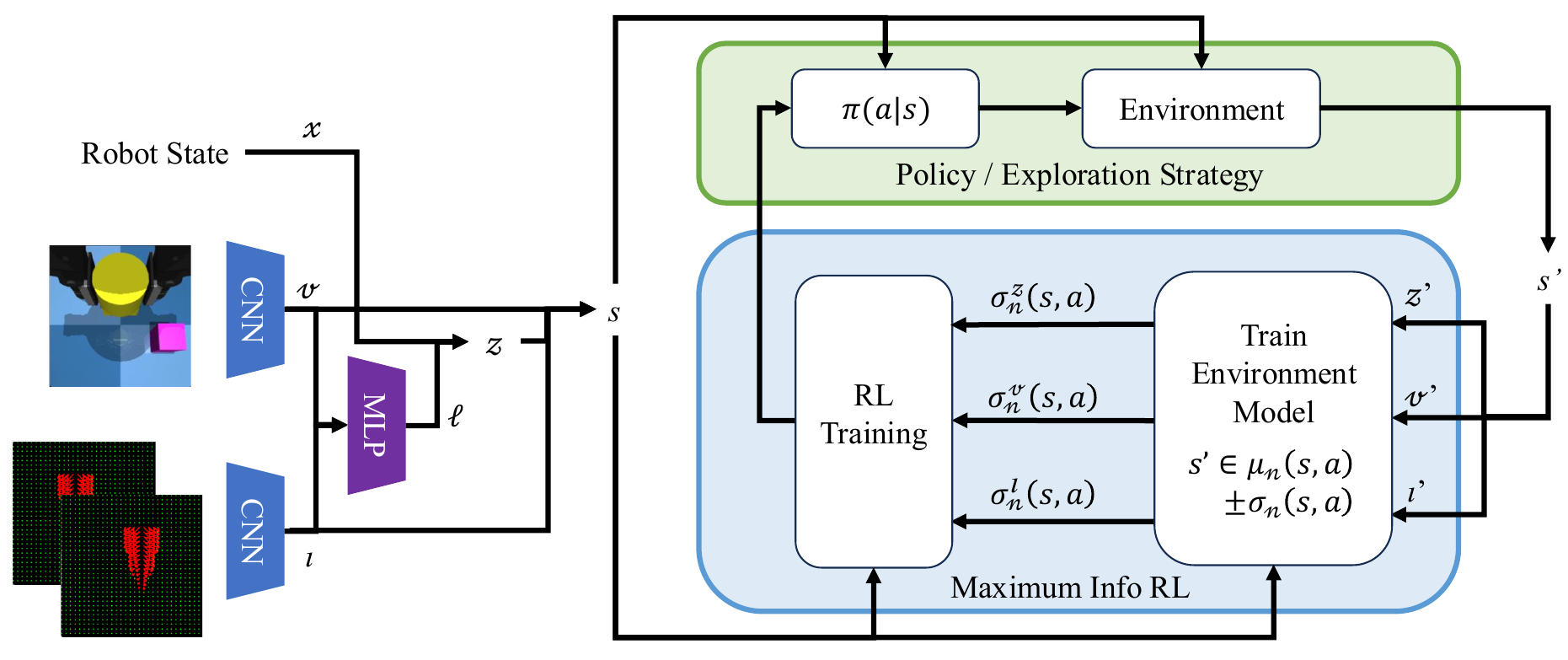}
    \caption{Overview of \alg. The state $s$ concatenates the fused latent $z$, a visual embedding $v$, and a tactile embedding $\iota$, where $v$ and $\iota$ are produced by separate CNN encoders from RGB and tactile force maps. At each step, the policy $\pi(a|s)$ acts in the environment, and the resulting transitions train an ensemble environment model that predicts the next state $s'$. The model's epistemic uncertainty decomposes per modality into $\sigma_n^z$, $\sigma_n^v$, and $\sigma_n^\iota$; unlike standard MaxInfoRL~\cite{sukhija2024maxinforl}, which rewards the agent for total predictive uncertainty, \alg explicitly weights the \emph{tactile} component $\sigma_n^\iota(s, a)$ in the exploration bonus. This anchors curiosity to the sense of touch, driving the agent toward contact-rich interactions and away from uncertainty in functionally irrelevant regions and transitions.}
    \label{fig:overview}
\end{figure}

Human babies learn to manipulate the world through touch~\cite{johansson2009coding}. Long before they can name an object or reason about its dynamics, they grasp, press, and reorient, and it is through these contacts that complex manipulation skills emerge. In fact, contact is the central piece of manipulation physics, and any agent that hopes to learn dexterous behavior must learn to seek out and understand contact.

By contrast, reinforcement learning typically discovers behavior by sampling actions isotropically. The consequence is an agent that allocates most of its training budget to motions in free space, far from the contacts that matter. In manipulation, the informative region of the state space is vanishingly small relative to its volume: a gripper a centimeter above an object and a gripper pressing into it occupy nearby states but induce entirely different dynamics, and only the latter teaches the agent anything about grasping. Random exploration is poorly equipped to discover this region, and once found, poorly rewarded to stay there.

Intrinsic motivation~\citep{aubret2019survey} methods based on forward dynamics disagreement or epistemic uncertainty improve substantially on undirected noise~\citep{sukhija2024maxinforl, pathak2017curiosity, sekar2020planning, sukhija2024optimistic, sukhija2025sombrl}, but their curiosity remains agnostic to contact. The agent is rewarded for resolving uncertainty wherever it happens to lie, and is just as content hovering a manipulator through unexplored free space as it is dwelling on a contact to explore its dynamics. Worse, because interaction signals are sparse and arise only upon contact, the states an agent most needs to be curious about are precisely the ones it visits least.

We argue that for robotic manipulation, tactile feedback is the natural prior for where curiosity should point. The agent should not seek one-off contact, nor should it seek uncertainty uniformly across the state space; it should seek uncertainty in the tactile channel, and across the contact modes that touch exposes. This reframes exploration around a simple inductive bias: the value of resolving uncertainty is not uniform across the senses, and the sense of touch is the one whose uncertainty most reliably indicates that something worth learning is taking place.

We introduce \alg{}, a framework that builds on this idea (\cref{fig:overview}). Rather than treating epistemic uncertainty as a single scalar over the full state, \alg{} decomposes it across sensory modalities and introduces a per-modality weighting on the resulting uncertainty terms. The key inductive bias is that the tactile component must be represented and weighted in the bonus, rather than, as in standard uncertainty-driven exploration, folding touch into an undifferentiated quantity where its contribution is diluted. With touch given non-negligible weight, the exploration bonus drives the agent toward contact-rich regions of the state space and rewards it for resolving the dynamics it finds there. The result is curiosity anchored to the sense of touch: the robot is drawn to contact, and it learns to grasp and manipulate objects without task rewards or expert demonstrations during exploration.

This formulation yields three contributions. First, we introduce a modality-weighted exploration objective, motivated by information-gain exploration, that explicitly prioritizes tactile predictive uncertainty. Second, a pure tactile-driven exploration phase learns emergent grasping behaviors entirely from scratch, and the interaction-dense dataset it collects supports offline learning of downstream pick-and-place policies without additional environment interaction. Third, the same principle makes the post-training of vision-language-action (VLA) models interaction-driven and sample-efficient: although these models are pre-trained without any tactile feedback, post-training with \alg{} substantially improves downstream performance while remaining highly sample-efficient.  We evaluate these claims in two simulation settings: reward-free exploration from scratch and sample-efficient post-training of frozen VLAs.

\section{Background}\label{sec: background}
In the following, we summarize a few key exploration strategies for RL and prior works that leverage tactile sensing for robot learning. A more detailed list of related work is also provided in Appendix \ref{sec: related_work}.
\subsection{The Exploration-Exploitation Dilemma in RL}
\looseness -1
The exploration-exploitation trade-off represents a crucial challenge in RL. Should the agent exploit its current knowledge to greedily maximize the rewards or explore new policies in the hope of discovering a better solution? Common exploration strategies in RL are $\epsilon$-greedy, Boltzmann exploration, and maximizing information gain. We discuss these in the following.

\paragraph{Problem Setting} \label{sec: problem setting}
We consider an infinite-horizon Markov decision process~\citep[MDP,][]{puterman2014markov}, defined by the tuple $(\gS, \gA, p, \gamma, r)$. The state and action spaces are assumed to be continuous, i.e., $\gS \subset \R^{d_s}, \gA \subset \R^{d_a}$, and the unknown transition kernel $p: \gS \times \gS \times \gA \to [0, \infty) $ represents the probability density of the next state $\vs_{t+1} \in \gS$ given the current state $\vs_{t} \in \gS$ and action $\va_{t} \in \gA$. At step $t$, the agent observes the state $\vs_t$ and samples an action $\va_t$ from the policy $\vpi: \gA \times \gS \to [0, \infty)$, $\va_t \sim \vpi(\va|\vs_t)$. Following the execution of the action $\va_t$ in the environment, the agent receives a reward $r: \gS \times \gA \times \gS \to \R_+$ and the goal of the agent is to  maximize the discounted sum of rewards:
 \begin{equation}
    \label{eq:rl-objective}
    \max_{\vpi \in \Pi} J(\vpi) = \max_{\vpi \in \Pi} \E_{\vs_0, \va_0, \dots} \left[\sum_{t = 0}^{\infty} \gamma^t r_t \right].
\end{equation}

\looseness=-1
The state-action critic $Q^{\vpi}$ and the value function $V^{\vpi}$ are defined as:
\begin{equation*}
    Q^{\vpi}(\vs_t, \va_t) = \E_{\vs_{t+1}, \va_{t+1} \sim \vpi, \dots}  \left[\sum_{l = 0}^{\infty} \gamma^l r_{t+l} \right], \;
    V^{\vpi}(\vs_t) = \E_{\va_t \sim \vpi, \vs_{t+1}, \va_{t+1} \sim \vpi, \dots}  \left[\sum_{l = 0}^{\infty} \gamma^l r_{t+l} \right].
    \label{eq: vf_defs}
\end{equation*}
\vspace{-3mm}
\paragraph{$\epsilon$--greedy exploration } \label{sec: bg_eps_greedy}
A common strategy to balance exploration and exploitation is $\epsilon$-greedy~\citep{kearns2002near, mnih2013playing, van2016deep}, where with probability $\epsilon$, the agent selects a random action $\va_t \sim \text{Unif}(\gA)$ to encourage exploration and otherwise acts greedily, i.e., $\va=\arg\max_{\va\in\gA} \; Q^{*}(\vs_t, \va)$. Here $Q^{*}$ is the estimate of the optimal state-action value function.  For continuous action spaces, \citet{lillicrap2015continuous, fujimoto2018addressing} use a deterministic policy $\vpi_{\theta}$, which maximizes the value function, in combination with random Gaussian noise to encourage exploration.

\paragraph{Boltzmann Exploration} Many RL algorithms are based on Boltzmann exploration~\citep{sutton2018reinforcement, szepesvari2022algorithms}. Here, the policy $\vpi$ follows the distribution below
\begin{equation}
    \vpi(\va|\vs) = Z^{-1}(\vs)\exp\left(\alpha^{-1}Q^{\vpi}(\vs, \va)\right),
   \label{eq: bolztman standard}
\end{equation}
where $Z^{-1}(\vs)$ is a normalization term, $\alpha$ is the temperature parameter that regulates exploration and $Q^{\vpi}$ is the soft-$Q$ function. \citet{haarnoja2018soft} show that \cref{eq: bolztman standard} corresponds to maximizing the entropy weighted reward $\tilde{r}(\vs, \va, \vs') =  r(\vs, \va, \vs') + \alpha \mathcal{H} \left(\vpi(\cdot\mid\vs)\right)$.

For $\alpha \to 0$, the policy acts greedily and maximizes  $Q^{\vpi}(\vs, \va)$, and for $\alpha \to \infty$ the policy explores uniformly, adding equal mass to all actions in $\gA$. Intuitively, Boltzmann exploration can be interpreted as a smoother alternative to $\epsilon$--greedy, with $\alpha$ serving a similar role to $\epsilon$ in controlling the degree of exploration.

Both $\epsilon$--greedy and Boltzmann exploration fail to account for the agent's ``lack of knowledge'' and do not encourage directed exploration strategic for the task. The agent explores by sampling random action sequences, which leads to suboptimal performance, particularly in challenging exploration tasks with continuous state-action spaces.

\paragraph{Maximizing Information Gain}
\looseness=-1
\citet{sukhija2024maxinforl} propose learning an uncertainty-aware dynamics model of the MDP and utilize the epistemic uncertainty or model disagreement $\vsigma_n( \vs, \va) \in \R_{+}^{d_{\vs}}$ as an intrinsic reward bonus for directing exploration. They propose the following modification to the Boltzmann distribution above
\begin{equation}
    \vpi(\va|\vs) \propto \exp\left(\alpha^{-1}\left(Q^{\vpi}(\vs, \va)  + \lambda \norm{\vsigma_n( \vs, \va)}\right)\right).
    \label{eq: bolztman max info}
\end{equation}
 Effectively, the agent explores by not only applying random actions but also visiting new state-action tuples that exhibit high epistemic uncertainty. This exploration strategy enables principled and directed exploration, enjoying strong theoretical guarantees on the agent's convergence~\citep{sukhija2024maxinforl, sukhija2025sombrl, bartos2026optimistic}. Similar to \cref{eq: bolztman standard}, \cref{eq: bolztman max info} can be viewed as maximizing the reward $\tilde{r}(\vs, \va, \vs') =  r(\vs, \va, \vs') + \alpha \mathcal{H} \left(\vpi(\cdot\mid\vs)\right) + \lambda \norm{\vsigma_n( \vs, \va)}$. \citet{sukhija2024maxinforl} also propose how to automatically select the temperature parameters $(\alpha, \lambda)$ and show that \cref{eq: bolztman max info} outperforms state-of-the-art RL algorithms for both state-based and visual control tasks.

\subsection{Tactile Sensing}
Tactile feedback can be available in many forms, varying widely in the richness of the signal they provide. At the lowest end of the spectrum are binary contact sensors~\cite{yin2023rotating}, which report only whether contact is present, and force-torque sensors, which measure aggregate wrench at a single point but discard any spatial information about where contact occurs. In this work, we focus on high-dimensional tactile feedback, which, much like human skin, resolves contact spatially and reports the distribution of pressure and shear across the sensing surface.

The most common forms of high-dimensional tactile feedback are tactile images~\cite{yuan2017gelsight} and force maps~\cite{sferrazza2019ground}, both of which are naturally interpretable as standard images and can therefore be processed with off-the-shelf vision architectures. Here we use force maps: a spatial binning of the contact patch in which each cell records the local pressure and shear forces, represented as a three-dimensional force vector. Because both representations share this image-like structure, force maps can be readily swapped for tactile images from vision-based sensors with little change to the rest of the pipeline, although its effectiveness with tactile images remains to be evaluated. We adopt force maps because they are convenient in practice: they are computable from real-world tactile sensors~\cite{sferrazza2019ground, sferrazza2022sim} and readily available in common simulation frameworks~\cite{sferrazza2023power, akinola2025tacsl}, which makes them well suited for studying tactile-driven exploration in both settings.

\subsection{Touch-guided exploration}
Touch has long been used for active information gathering. \citet{javdani2013efficient} select touch actions to reduce uncertainty over object pose using submodular objectives, and in \cite{javdani2014near}, study touch-based localization, where sensing need only resolve uncertainty that affects the subsequent decision.

Within reinforcement learning, tactile feedback has also been used to guide exploration. \citet{vulin2021improved} reward crossing a cumulative contact-force threshold and prioritize contact-rich replay; \citet{huang2019learning} combine impact penalties with surprise from an impact-penalty predictor; and \citet{rajeswar2021touchbased} combine visual-to-haptic prediction error with visual forward-model error. Our contribution is to explicitly weight modality-specific ensemble disagreement, including uncertainty over spatial tactile representations, and study its effect on exploration and downstream learning. Spatial normal/shear force maps retain contact-location and force-distribution information lost by aggregation.

\section{\alg: Tactile Exploration Drives Efficient Learning}

A key limitation of the objective in \cref{eq: bolztman max info} is that it aggregates predictive uncertainty across the state representation without explicitly controlling each modality’s contribution. Since $\vsigma_n(\vs,\va) \in \mathbb{R}_{+}^{d_{\vs}}$, maximizing the norm $\norm{\vsigma_n(\vs,\va)}$ implicitly encourages exploration uniformly in every direction of the state space. However, in many robotic systems, effective exploration is modality-dependent and may be concentrated in only a subset of the state space.

For instance, in robotic manipulation tasks, exploration of states associated with physical interaction between the manipulator and the object is substantially more informative than exploration in regions where the manipulator remains in free space. To incorporate this inductive bias, we augment the state representation with tactile information and introduce a weighting vector that prioritizes uncertainty in task-relevant dimensions. In particular, let $\vv \in \R^{d_{\vv}}$ be the visual state representation of the robot extracted from images, $\viota \in \R^{d_{\viota}}$ the tactile state representation extracted from the tactile sensors, and $\vz \in \R^{d_{\vz}}$ a latent representation of the robot's state. The full state is defined as $\vs = [\vz^\top, \viota^\top, \vv^\top]^\top$. We then decompose epistemic uncertainty across modalities and modify the exploration strategy from \cref{eq: bolztman max info} as follows:
\begin{equation}
    \vpi(\va|\vs) \propto \exp\left(
    \alpha^{-1} \left(Q^{\vpi}(\vs,\va)
    +
    \lambda\left(\omega_{\vv} \norm{\vsigma^{\vv}_n(\vs,\va)} + \omega_{\viota} \norm{\vsigma^{\viota}_n(\vs,\va)} + \omega_{\vz} \norm{\vsigma^{\vz}_n(\vs,\va)}\right)\right)
    \right).
    \label{eq: bolztman max info adapted}
\end{equation}

Here, $\vsigma^k_n(\vs,\va)$ denotes modality-specific epistemic uncertainty, and $\omega_\vk \ge 0$ controls the relative contribution of each sensing channel. Crucially, this formulation enables tactile-centric exploration, which can be enforced by assigning dominant weight to $\viota$. For instance, setting $\omega_{\viota} \gg \omega_{\vv}, \omega_{\vz}$ biases exploration toward contact-rich regions, and $\omega_{\vv} = \omega_{\vz} = 0$ yields purely tactile-driven information gain. Changing these weights changes the exploration bonus, rather than specifying which observations the policy receives. Purely tactile weighting is not assumed to be optimal for every task, since visual or latent disagreement can encourage complementary exploration.

\section{Experiments}
\label{sec:exp}

We evaluate \alg in two simulation settings. The central hypothesis behind \alg is that tactile feedback provides a crucial signal for enabling meaningful and directed exploration. To investigate this hypothesis, we study whether touch encourages contact-rich exploration in simulation. In reward-free exploration, an intrinsically motivated policy collects data without task rewards; downstream policies are then trained offline on the relabeled, fixed dataset. In diffusion steering, the steering policy is trained online using task rewards augmented with the intrinsic bonus, while the pretrained VLA remains frozen. We report means and standard errors over five seeds for reward-free exploration and ten seeds for diffusion steering.

\subsection{Tactile-Driven Reward-Free Exploration}

To isolate the role of the exploration bonus, we consider reward-free data collection with $r\equiv 0$. The agent optimizes only the uncertainty-driven intrinsic reward defined from \cref{eq: bolztman max info adapted}, without access to task rewards. All modality-weighted exploration runs use zero extrinsic reward. We systematically ablate different weighting configurations $(\omega_{\vv}, \omega_{\viota}, \omega_{\vz})$ in \cref{eq: bolztman max info adapted} to analyze the relative contribution of visual, tactile, and latent uncertainty to exploration behavior. We also include a random-exploration baseline, which collects data using random actions without guidance from task rewards or model disagreement.

\begin{figure}
    \centering
    \includegraphics[width=\linewidth]{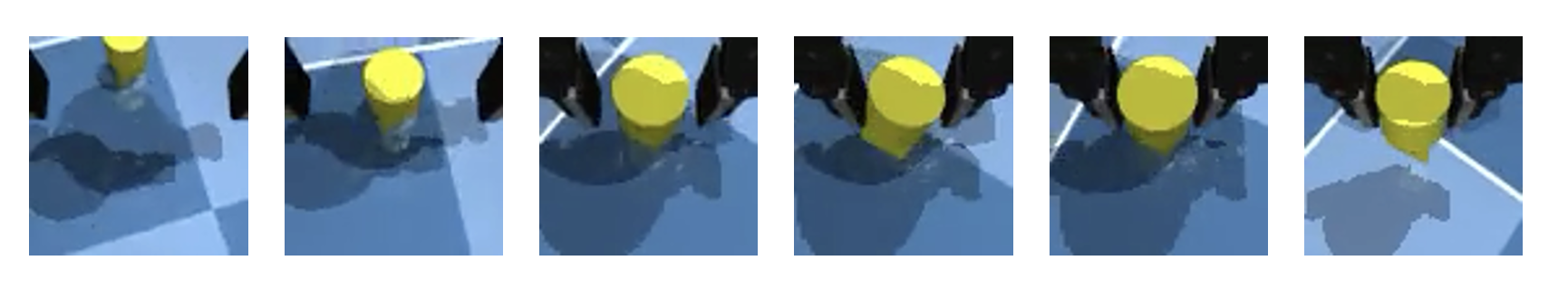}
    \includegraphics[width=.9\linewidth]{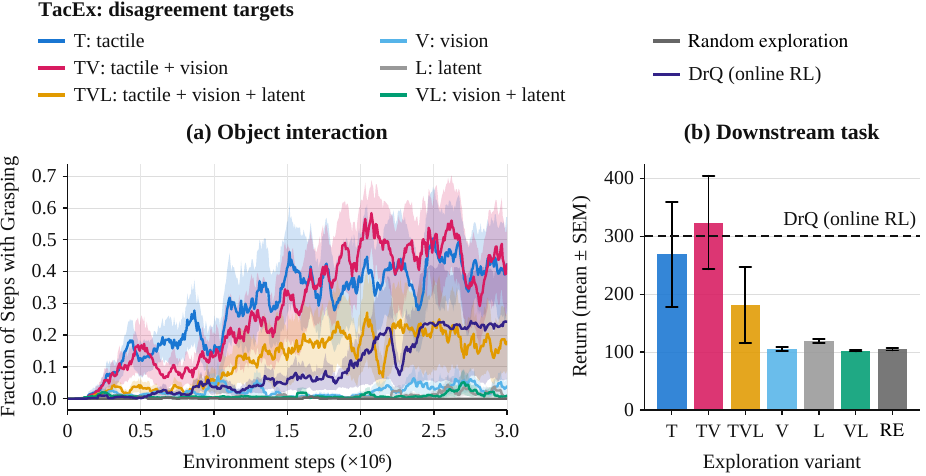}
    \caption{Tactile-driven exploration with a single object. \textit{Top:} selected exploration rollout frames, illustrating approach, grasp, and lift. \textit{Bottom:} (a) fraction of steps spent grasping during exploration; (b) downstream pick-and-place return after offline policy learning on the collected dataset. We compare six disagreement-based exploration variants with random exploration, all without task rewards during data collection. Downstream returns are averaged over the final ten evaluation checkpoints within each seed. DrQ is a separate online RL baseline trained directly on the downstream task reward; the dashed line in (b) denotes its evaluation return. We report means and standard errors across five seeds.}
    \label{fig: single object touch}
\end{figure}
\paragraph{Does touch lead to contact-rich and meaningful exploration?} In \cref{fig: single object touch}, we evaluate \alg in a simulated environment with a robotic gripper and a single object \cite{sferrazza2023power}. Successful interaction with the object depends entirely on the agent’s exploration behavior. This setting allows us to vary the exploration bonus while keeping the manipulation scene fixed. During the exploration phase, the agent receives no extrinsic reward $r$, and exploration is therefore driven solely by the uncertainty-based intrinsic objective in \cref{eq: bolztman max info adapted}. We compare several configurations of the weighting vector, namely: (i) tactile-only exploration $(\omega_{\vv}=0, \omega_{\viota}=1, \omega_{\vz}=0)$, (ii) combined tactile and visual exploration $(\omega_{\vv}=1, \omega_{\viota}=1, \omega_{\vz}=0)$, (iii) vision-only exploration $(\omega_{\vv}=1, \omega_{\viota}=0, \omega_{\vz}=0)$, and (iv) latent-state-only exploration $(\omega_{\vv}=0, \omega_{\viota}=0, \omega_{\vz}=1)$.

To further assess the benefits of contact-rich exploration, we additionally consider a downstream sparse-reward pick-and-place task, where the objective is to grasp the object and lift it to a target location. The reward uses a finite-support tolerance margin around the spatial target, following the DeepMind Control Suite~\citep{tassa2018deepmind}. After reward-free collection, we freeze the replay buffer, relabel its transitions with this reward, and train SAC entirely offline without collecting additional training transitions.  Importantly, during data collection, the agent is driven purely by the uncertainty-based exploration objective, without access to task rewards. Finally, we compare against an online RL baseline, DrQ~\cite{kostrikov}, trained directly on the sparse-reward task.

From \cref{fig: single object touch}, we draw several conclusions. First, incorporating tactile-driven disagreement increases the frequency with which the agent interacts with and attempts to grasp the object. In contrast, without tactile disagreement, the agent rarely engages with the object at all, suggesting that uniform uncertainty over the state space does not necessarily translate into meaningful exploration. This highlights the role of tactile sensing as an effective inductive bias that promotes contact-rich behavior in manipulation tasks.

Interestingly, variants that prioritize tactile disagreement also induce substantially more object interaction than the sparse-reward online RL baseline and may yield higher downstream pick-and-place returns after offline learning on their collected datasets. Together with the increased grasping fraction, this suggests that guiding exploration toward physically informative, contact-based experiences is critical for learning effective manipulation policies.

\begin{figure}
    \centering
    \includegraphics[width=\linewidth]{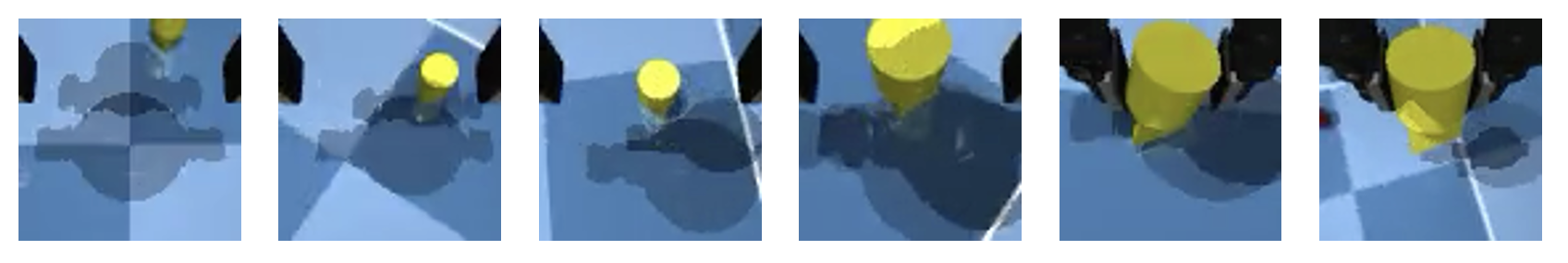}
    \includegraphics[width=\linewidth]{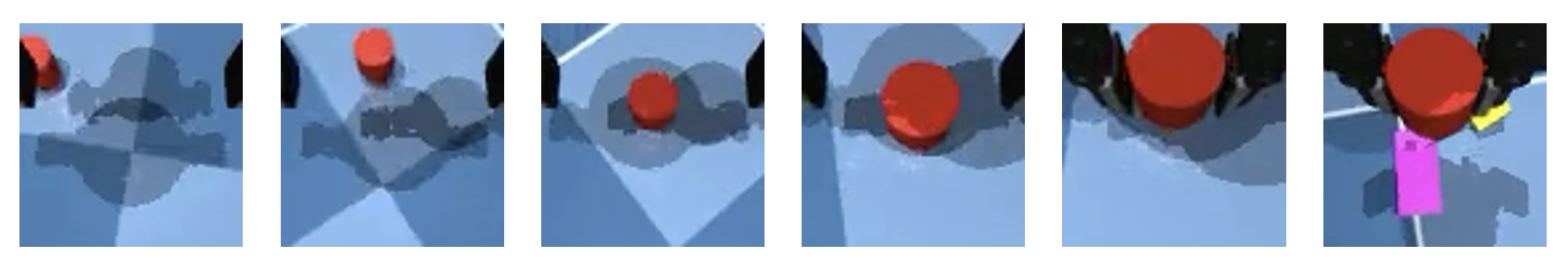}
    \includegraphics[width=\linewidth]{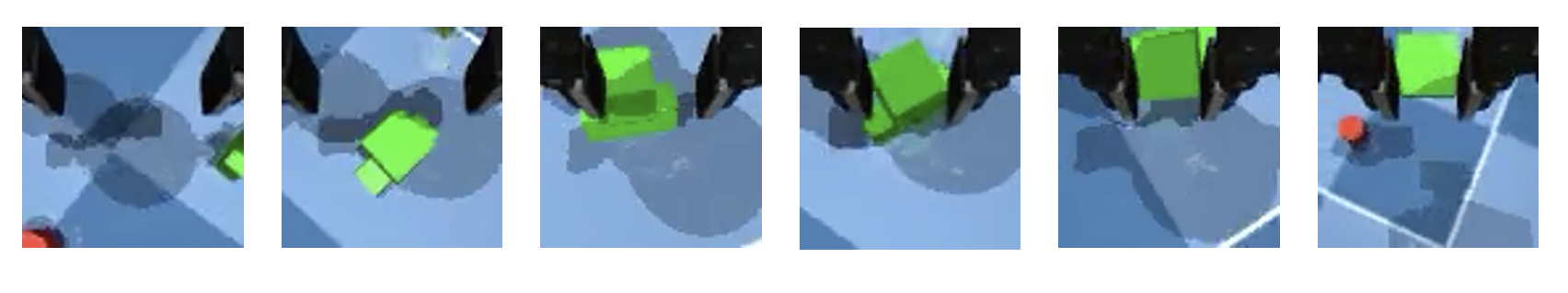}
    \includegraphics[width=\linewidth]{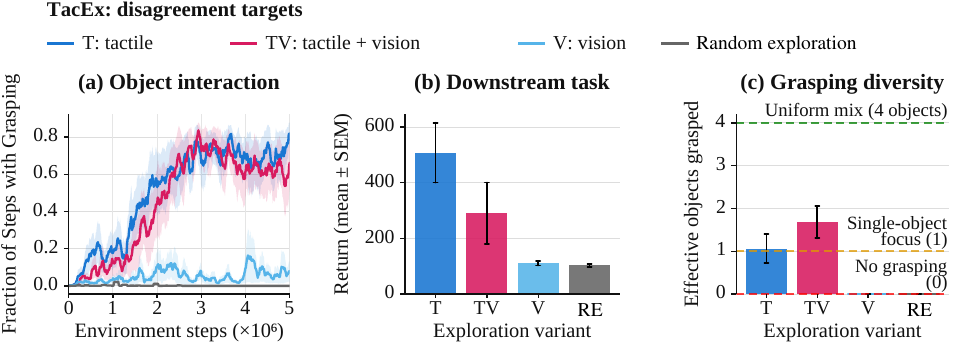}
    \caption{Tactile-driven exploration with multiple objects. We extend the setup of \cref{fig: single object touch} to four objects. \textit{Top:} selected exploration rollout frames. Rows illustrate grasping different objects, with time progressing from left to right. \textit{Bottom:} (a) fraction of steps spent grasping; (b) downstream pick-and-place return after offline policy learning; (c) grasping diversity, measured as the effective number of objects grasped, $2^H$. Downstream returns are averaged over the final ten evaluation checkpoints within each seed. We report means and standard errors across five seeds.}
    \label{fig: second object touch}
\end{figure}

Finally, we investigate how the exploration agent behaves in a more complex environment with multiple objects. To this end, we introduce three additional objects, resulting in four objects in total, and report the corresponding results in \cref{fig: second object touch}. To analyze how the agent distributes its interactions across these objects, we quantify the effective number of objects being grasped during exploration. Specifically, we compute the entropy $H = -\sum_{i=1}^{4} p_i \log_{2} p_i$, where $p_i$ denotes the proportion of grasps attributed to object $i$, and report the effective number of objects as $2^H$.

From \cref{fig: second object touch}, we further observe that tactile disagreement plays a crucial role in enabling contact-rich manipulation and strong downstream pick-and-place performance. Interestingly, augmenting tactile disagreement with visual disagreement increases exploration diversity, leading the agent to interact with multiple objects (effective number of objects $\approx 2$), whereas purely tactile-driven exploration results in more focused behavior, with an effective number of objects close to 1. While tactile disagreement yields slightly better performance on the downstream pick-and-place task—where only a single object must be manipulated—incorporating visual disagreement encourages broader exploration and maintains curiosity across multiple objects in the scene.

\subsection{Tactile Diffusion Steering}
\label{sec}
The previous experiments show that tactile uncertainty can guide exploration when learning from scratch. We now ask whether the same idea can improve larger pretrained policies without updating them. We use diffusion steering~\citep{wagenmaker2025steering}, where a lightweight MLP predicts a noise action $\vw_t \in \gW$ and a frozen VLA policy $\pi_0$ maps this noise to robot actions. We build on the LIBERO-90 suite~\citep{liu2023libero} and add tactile sensing~\citep{sferrazza2023power} to the gripper.

The base policy $\pi_0$ receives the same visual observations, proprioceptive state, and language command as during pre-training. In the tactile variants, the learned noise-prediction policy and critics additionally receive tactile observations $\viota_t$ from the gripper and for \alg the uncertainty-based exploration reward from  \cref{eq: bolztman max info adapted} is used to direct exploration.

\paragraph{Baseline and method}
\algdsrlsac trains an MLP policy $\pi^\vw$ with SAC to predict the latent diffusion noise $\vw_t$ from the current learner state $\vs_t$, while the frozen VLA policy $\pi_0$ maps this noise to robot actions. This baseline uses visual and proprioceptive observations, but no tactile input, and optimizes only the sparse task reward.

\algdsrltacmaxinfosac augments the learner state with tactile observations $\viota_t$ from the gripper and adds the intrinsic reward from \cref{eq: bolztman max info adapted}. We train an ensemble dynamics model on online transitions $(\vs_t,\vw_t,r_t,\vs_{t+1})$, using the current learner state and predicted noise action as inputs. The ensemble predicts the next transition in selected observation modalities and the task reward; disagreement between ensemble heads serves as an epistemic uncertainty estimate. The policy is therefore encouraged to choose latent noise actions whose outcomes are informative under the learned latent-noise MDP. In the main experiments, we evaluate tactile-only disagreement, $(\omega_\vv,\omega_{\viota},\omega_\vz)=(0,1,0)$, and two mixed variants, $(1,1,0)$ and $(0,1,1)$.

\paragraph{Experimental setup}
We evaluate on eight contact-rich tasks from the LIBERO-90 suite. These tasks require the robot to establish and exploit physical contact with objects, containers, or articulated scene elements, making them a natural testbed for tactile diffusion steering. All methods use the same pretrained $\pi_0$ checkpoint, the same language command for each task, and the same online interaction budget within each task. We report success rate, shown as average evaluation return, as a function of online environment interaction and average over random seeds.

\begin{figure*}[t]
    \centering
    \includegraphics[width=\linewidth]{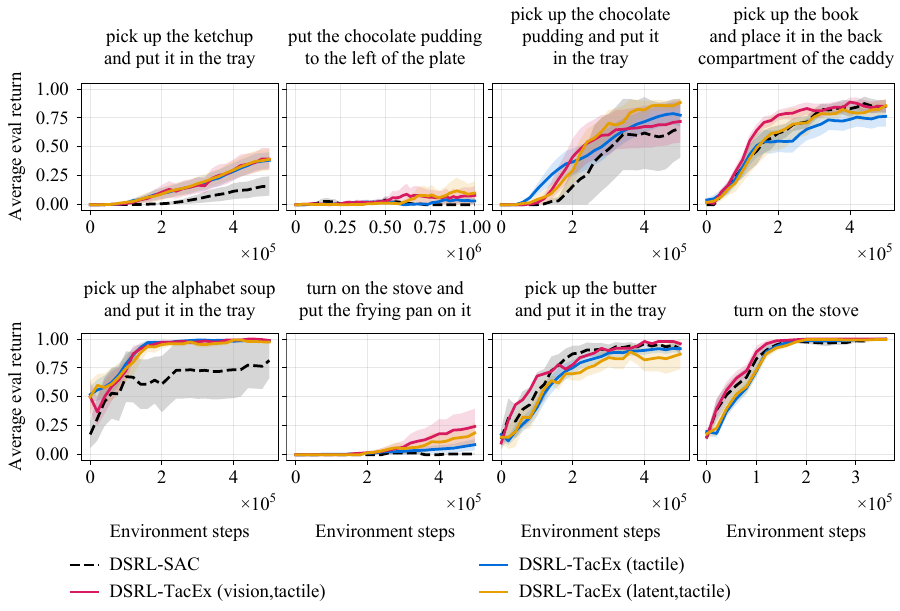}
    \caption{Online diffusion-steering performance on contact-rich LIBERO-90 tasks. \algdsrlsac uses visual and proprioceptive state only and optimizes sparse task reward with SAC. \algdsrltacmaxinfosac additionally receives tactile observations and uses ensemble disagreement weightings $(\omega_\vv,\omega_{\viota},\omega_\vz)$ over tactile $(0,1,0)$, vision+tactile $(1,1,0)$, or latent+tactile $(0,1,1)$ predictions as an intrinsic exploration bonus. Shaded regions denote standard error across 10 random seeds.}
    \label{fig:dsrl_all_envs}
\end{figure*}

\paragraph{Results}
\Cref{fig:dsrl_all_envs} shows that tactile MaxInfo diffusion steering can improve over \algdsrlsac across the evaluated contact-rich tasks. The effect is task dependent: when the pretrained $\pi_0$ prior already explores sufficiently, additional intrinsic reward is less important and can sometimes be distracting, which may reflect differences in the pretrained policy and the interactions required for completion. Still, across the selected tasks, the best-performing curves typically include tactile disagreement. These results suggest that touch is useful not only as an observation modality, but also as a way to shape exploration during post-training of frozen VLAs.

\pagebreak
\paragraph{Isolating the role of tactile uncertainty}

In Appendix \ref{app:dsrl_ablations}, we ablate the effect of adding touch purely as a state modality versus using it to direct exploration. There, \cref{fig:dsrl_touch_ablations} separates tactile observations, information-gain exploration, and tactile disagreement in the bonus. On this task, tactile observations improve performance, information-gain exploration improves over SAC, and the highest-performing exploration variants also include tactile disagreement.

The five-task comparison in \cref{fig:dsrl_extended_comparison} of Appendix \ref{app:dsrl_ablations} extends this analysis. We compare tactile-only, tactile+vision, and tactile+latent bonuses with bonuses computed over visual, latent, or state predictions without an explicit tactile target. Tactile+vision achieves the highest reported mean final return on all five tasks, while other prediction targets remain competitive on several tasks. Together, these results support tactile uncertainty as a useful, task-dependent exploration signal, with visual uncertainty providing complementary benefits. Additional baselines in \cref{fig:dsrl_extended_comparison} of Appendix \ref{app:dsrl_ablations} examine the tactile representation and the post-training algorithm. Two summed-taxel controls use aggregate touch either directly as an intrinsic reward or as a prediction target for ensemble disagreement, testing the value of retaining spatial tactile information. DSRL-PPO replaces SAC in the noise-steering policy, while DPPO~\citep{ren2024dppo} fine-tunes the diffusion policy itself. Across the five evaluated tasks, the tactile+vision variant achieves higher mean final return than both summed-taxel controls and the PPO-based baselines.

\section{Conclusion}

In this work, we introduced \alg, a framework that establishes tactile feedback as a core prior for directed exploration in robotic manipulation. By prioritizing tactile uncertainty, \alg avoids the inefficiency of traditional reinforcement learning methods that spend large portions of training on uninformative free-space exploration. Our results show that tactile-driven curiosity enables the discovery of meaningful contact dynamics and emergent grasping behaviors without external rewards or demonstrations, and that the interaction-rich data collected through \alg supports offline learning of downstream pick-and-place policies without additional environment interaction. Furthermore, \alg improves the sample efficiency of post-training Vision-Language-Action (VLA) models, despite the VLA itself being pre-trained with datasets without tactile observations. Finally, we find that combining tactile and visual uncertainty encourages both focused contact exploration and broader interaction diversity. Overall, these findings demonstrate that touch is not merely an observation modality, but a critical signal for efficient exploration and skill acquisition in robotic manipulation.

\subsection*{Limitations and future work} \label{ssec:lim_future}
Our study has several limitations that open avenues for future work. First, while we adopt force maps precisely because they transfer readily between simulated and real tactile sensors~\cite{sferrazza2022sim}, validating tactile-driven exploration on physical hardware---where contact is noisier and resets are costly---remains an important next step. We do not evaluate sensitivity to tactile noise, calibration drift, or sensor wear, and the present experiments do not establish whether these effects can be separated from uncertainty about contact dynamics. Moreover, contact-seeking curiosity does not itself enforce safe contact forces or prevent unintended interactions with distractor objects. Offline initialization and explicit safety constraints are possible directions for physical deployment, but their effectiveness with \alg remains to be evaluated.

Second, our experiments use a single parallel-jaw gripper and a single tactile representation; whether tactile curiosity scales to dexterous multi-fingered hands, richer contact modes, and vision-based tactile images is left open, though our pipeline is designed to accommodate such sensors with minimal modification. The evaluated tasks do not test fragile or deformable objects, or establish force regulation in tightly constrained manipulation. Third, the per-modality weights $(\omega_v, \omega_\iota, \omega_z)$ are fixed hyperparameters. Our multi-object results suggest a tradeoff between focused contact exploration and interaction diversity that these weights control, and learning or adapting them online is a natural direction. Finally, while we show that tactile uncertainty improves the post-training of frozen VLAs, we do not explore whether the contact-rich data collected through \alg could be fed back into pre-training, closing the loop between tactile exploration and large-scale policy learning.


\pagebreak
\acknowledgments{We thank Manuel Wendl, Jasmine Bayrooti, Yarden As, and Bruce D.~Lee for their valuable feedback on this work. This work was supported as a part of NCCR Automation, a National Centre of Competence in Research, funded by the Swiss National Science Foundation (grant number 51NF40\_225155). This work was also funded by ONR MURI N00014-22-1-2773. 
B.~Sukhija was supported by ELSA (European Lighthouse on Secure and Safe AI) funded by the European Union under grant agreement No. 101070617. P. Abbeel holds concurrent appointments as a Professor at UC Berkeley and as an Amazon Scholar. This paper describes work performed at UC Berkeley and is not associated with Amazon.

Numerical simulations were performed on the ETH Zürich Euler cluster. Parts of this text were revised with the assistance of a large language model to assist with writing and editing as well as for code development throughout this work; the authors remain responsible for all content.}


\bibliography{references}  
\clearpage


\appendix
\section{Extended Related Work} \label{sec: related_work}
 \paragraph{Learning to Manipulate With Touch}

 A growing body of work leverages tactile sensing for robotic manipulation~\citep{tian2019manipulation, dong2021tactile, lee2024dextouch, xu2024tactilebased}. Prior works have shown that combining vision and touch can improve manipulation performance relative to using visual observations alone~\citep{Calandra_2018, sferrazza2023power}. Beyond its role as an observation modality, tactile feedback has also been used explicitly to guide exploration during learning. Prior methods reward force magnitude \citep{vulin2021improved} instead of epistemic uncertainty, curiosity over a scalar impact-force predictor \citep{huang2019learning}, or cross-modal prediction error that can conflate reducible epistemic with aleatoric uncertainty and uses aggregate force/torque with touch in the reward and state \citep{rajeswar2021touchbased}. \alg instead rewards modality-specific epistemic uncertainty, which decays as dynamics are learned, and uses spatial normal/shear force maps that preserve contact location, multiple contacts, rotation, and slip cues. Since tactile signals are inherently sparse and only available upon contact, exploration and learning with touch are especially challenging.

Because informative contact signals arise only when the robot interacts with its surroundings, collecting useful tactile experience remains an exploration challenge. In this work, we study this problem and investigate how tactile feedback can be used to direct exploration in reinforcement learning. More specifically, we investigate the benefits of touch-driven exploration in visually guided manipulation tasks.

 \paragraph{Exploration Strategies in Reinforcement Learning}
 Efficient exploration remains one of the central challenges in reinforcement learning, particularly in high-dimensional continuous control problems. Classical approaches such as $\epsilon$--greedy exploration~\citep{kearns2002near, mnih2013playing, van2016deep} and Boltzmann exploration~\citep{sutton2018reinforcement, szepesvari2022algorithms} encourage exploration through randomized action selection or entropy maximization. While effective in simple settings, these methods explore indiscriminately and fail to account for the agent's lack of knowledge about the environment. To address this limitation, several works propose uncertainty-aware exploration strategies that explicitly direct exploration toward poorly understood regions of the state-action space~\citep{sukhija2024maxinforl, sukhija2025sombrl,pathak2017curiosity, burda2018exploration}.

 In particular, \citet{sukhija2024maxinforl} formulate exploration as maximizing information gain by augmenting the policy objective with epistemic uncertainty estimates derived from learned dynamics models. Such approaches have demonstrated strong empirical performance~\cite{zheng2025learningsoftroboticdynamics, iten2026sampleefficientscalableexplorationcontinuoustime, iten2026modelbasedreinforcementlearningcontrol} and improved sample efficiency in continuous control and visual RL tasks. An aggregate uncertainty objective, however, does not explicitly control the contributions of different sensory modalities. In robotic manipulation, informative interactions are often concentrated around contact-rich regions of the state space, suggesting that exploration should preferentially focus on tactile interaction. Our work builds on uncertainty-driven exploration and introduces a modality-aware exploration bonus that weights tactile uncertainty as a distinct term to guide exploration more effectively in manipulation tasks.

 \paragraph{Role of Touch in Post-Training of VLAs}
 Vision-language-action (VLA) models, such as $\pi_0$~\citep{black2026pi0visionlanguageactionflowmodel}, can leverage large robot demonstration datasets that do not include tactile observations. Incorporating tactile feedback into such models at scale remains challenging, since it requires additional sensing hardware, calibration, and large multimodal datasets with aligned tactile observations and robot actions. FuSe~\citep{jones24fuse} addresses this challenge by finetuning pretrained generalist robot policies on smaller heterogeneous-sensor datasets, using language as a common grounding space between modalities such as vision and touch. Tactile-VLA~\cite{huang2025tactile}  incorporates tactile inputs into the VLA backbone and finetunes the model end-to-end to predict both motion and contact-force targets. In contrast, we ask whether touch can improve RL-based post-training while keeping the pretrained VLA and its input modalities unchanged.

 We study this question in the diffusion-steering~\cite{wagenmaker2025steering} setting. Rather than finetuning the pretrained VLA, we keep it fixed and train a lightweight MLP policy that predicts the diffusion noise used by the frozen policy. The resulting policy chain is $\vs_t \xrightarrow{\pi^\vw} \vw_t \xrightarrow{\pi_0} \va_t$, where $\pi^\vw$ predicts the latent-noise action $\vw_t$ and the frozen VLA $\pi_0$ maps this noise to the robot action $\va_t$, conditioned on its original visual, proprioceptive, and language inputs.

 We show that tactile feedback can help this post-training procedure in two complementary ways: first, by enriching the state representation available to the noise-prediction policy, and second, by providing an exploration signal through ensemble disagreement over tactile predictions. Concretely, we train a tactile diffusion-steering policy using the modality-weighted MaxInfoRL objective from \cref{eq: bolztman max info adapted}, and evaluate whether incorporating tactile observations and tactile disagreement improves performance on contact-rich tasks.

\section{Implementation Details and Hyperparameters}
\label{app:hyperparameters}

We use the same SAC and MaxInfoSAC hyperparameter choices as \cite{sukhija2024maxinforl,wagenmaker2025steering} unless explicitly noted. Within the SAC/MaxInfoSAC modality ablations, we hold the optimization settings, model capacity, update schedule, and evaluation protocol fixed while varying tactile observations and disagreement targets. This matching does not extend to PPO and DPPO, which use different optimization proce- dures. This makes the ablations isolate the role of tactile sensing and tactile uncertainty as outlined in Appendix \ref{app:dsrl_ablations}. Hyperparameters for the LIBERO experiments using DSRL are outlined in \cref{tab:appendix-hparams}.

\begin{table*}[t]
    \centering
    \small
    \caption{Main hyperparameters used for the LIBERO simulation experiments. Values are shared by the SAC baseline and MaxInfo-SAC ablations unless marked otherwise.}
    \label{tab:exp_details}
    \label{tab:appendix-hparams}
    \begin{tabular}{p{0.36\linewidth}p{0.56\linewidth}}
        \toprule
        Hyperparameter & Value \\
        \midrule
        \multicolumn{2}{l}{\textit{Environment}} \\
        Suite / task & LIBERO-90, task IDs 58, 69, 61, 77, 55, 21, 56, 20 \\
        Robot gripper & PandaGripper from \cite{sferrazza2023power}\\
        \midrule
        \multicolumn{2}{l}{\textit{Training}} \\
        Environment steps & $5 \times 10^5$ to $1 \times 10^6$ (task 69) \\
        Batch size & 256 \\
        Discount factor & 0.999 \\
        Update-to-data ratio & 20 gradient steps per environment step \\
        Online update warmup & 500 steps \\
        Policy query frequency & 20 environment steps \\
        Action magnitude & 1.0 \\
        \midrule
        \multicolumn{2}{l}{\textit{Observations}} \\
        Image resolution & $64 \times 64$ \\
        Tactile observation shape & $32 \times 64 \times 3$ \\
        \midrule
        \multicolumn{2}{l}{\textit{SAC backbone}} \\
        Actor / critic learning rate & $1 \times 10^{-4}$ / $3 \times 10^{-4}$ \\
        Temperature learning rate & $3 \times 10^{-4}$, automatic target entropy \\
        Target-network smoothing & $\tau = 0.005$ \\
        Actor / critic MLP & $(128,128,128)$ \\
        Visual encoder & CNN with channels $(32,32,32,32)$, latent dimension 50 \\
        Tactile encoder MLP & $(256,256)$ \\
        \midrule
        \multicolumn{2}{l}{\textit{MaxInfo-SAC}} \\
        Dynamics-entropy ($\lambda$) learning rate & $3 \times 10^{-4}$ \\
        Initial dynamics-entropy ($\lambda$) & 1.0 \\
        Dynamics model learning rate & $1 \times 10^{-3}$ \\
        Dynamics model weight decay &$1 \times 10^{-4}$ \\
        Dynamics model architecture & $5\times(256,256)$ \\
        \bottomrule
    \end{tabular}
\end{table*}

\section{Ablations}
\label{app:dsrl_ablations}

We compare four diffusion-steering variants. \algdsrlsac is the standard diffusion-steering baseline: an MLP policy $\pi^\vw$ is trained with SAC to predict the latent noise $\vw_t$ from the current state $\vs_t$, while the frozen VLA policy $\pi_0$ maps this noise to actions. This method uses visual and proprioceptive observations, but no tactile input. \algdsrltacsac uses the same SAC learner in the latent-noise action space, but augments the learner state with tactile observations $\viota_t$ from the gripper. This baseline tests whether tactile information improves diffusion steering as an additional observation modality, without changing the exploration objective.

We then add exploration that explicitly seeks informative transitions in the latent-noise MDP. For this, \algdsrlmaxinfosac and \algdsrltacmaxinfosac train an uncertainty-aware statistical dynamics model on the online replay buffer $(\vs_t,\vw_t,r_t,\vs_{t+1})$. The model takes the current learner observation and the predicted noise action $\vw_t$ as input, and predicts the resulting transition in selected observation modalities, as well as the task reward. We represent this model as an ensemble, so that epistemic uncertainty can be estimated from the disagreement between ensemble members~\cite{pathak2019self}. Intuitively, if different ensemble heads make different predictions about what will happen after applying a noise action $\vw_t$, then executing this noise action is expected to provide information about the latent-noise MDP.

\begin{figure*}[t]
    \centering
    \includegraphics[width=\linewidth]{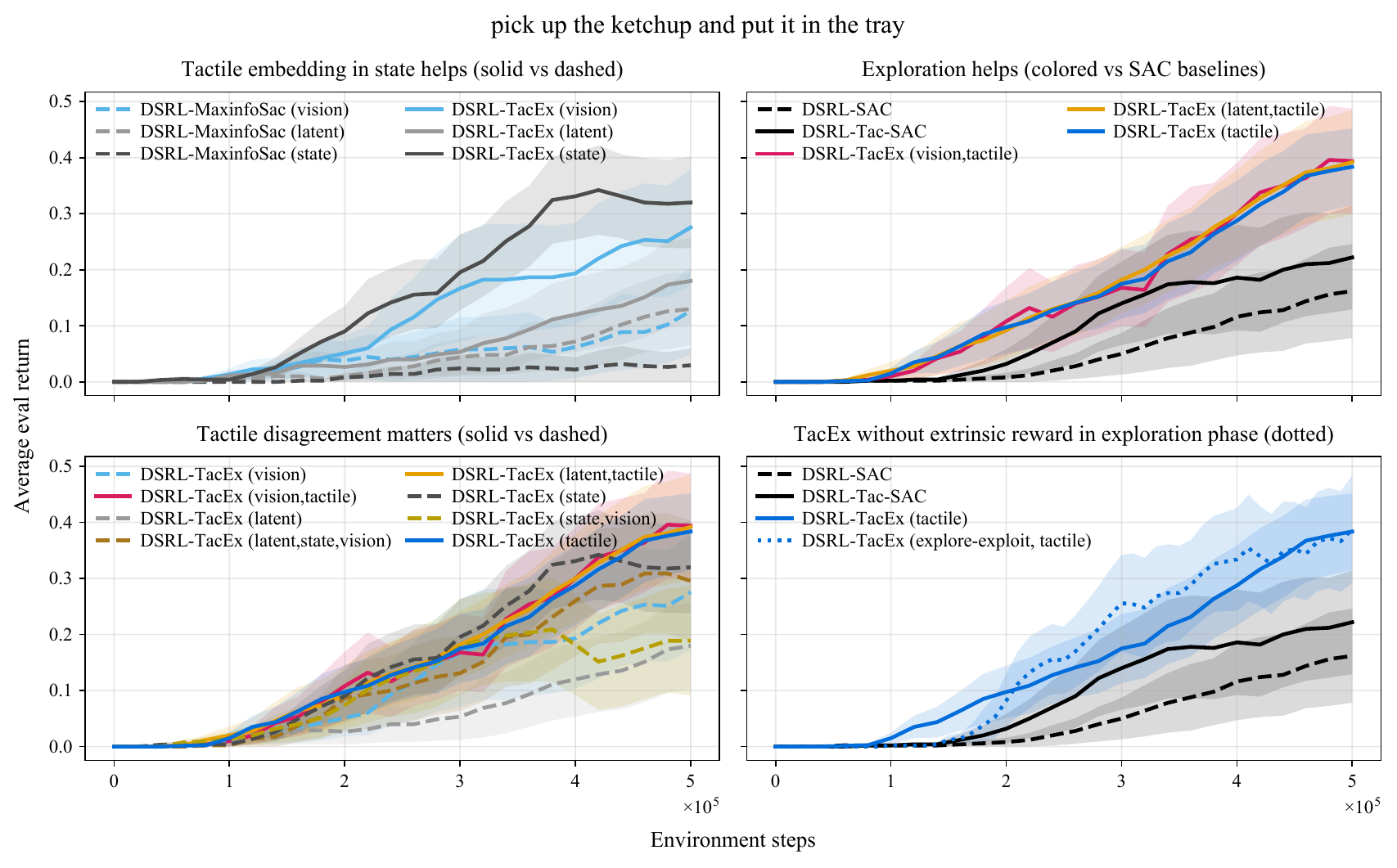}
    \caption{Ablations on LIBERO task 58. \textit{Top row:} solid curves use tactile observations in the learner state, dashed curves remove tactile observations from the learner state. \textit{Bottom-left:} dashed curves remove tactile modalities from the intrinsic reward. \textit{Bottom-right:} the dotted curve shows the explore-then-exploit schedule for comparison. In the \textit{right-hand} panels, black curves denote SAC diffusion-steering baselines, while colored curves add \alg-style exploration. Adding tactile observations improves diffusion steering, but the best-performing exploration variants are those that also include tactile prediction disagreement in the intrinsic bonus. We report the average evaluation return over 10 random seeds along with standard error bands.}
    \label{fig:dsrl_touch_ablations}
\end{figure*}

\algdsrlsac trains the noise-prediction policy using only the sparse task reward. In contrast, \algdsrlmaxinfosac augments this reward with an intrinsic bonus proportional to ensemble disagreement, encouraging the policy to choose noise actions whose outcomes are informative under the learned latent-noise dynamics model. In \algdsrlmaxinfosac, this disagreement is computed only over non-tactile modalities, i.e., $\omega_{\viota}=0$ since tactile observations are not present. \algdsrltacmaxinfosac uses a similar maximum information objective, but additionally receives tactile observations whose embeddings $\viota$ are included in the state $\vs$ and may include the tactile modality when computing disagreement with different weights. This tests whether tactile uncertainty provides a useful exploration signal for discovering contact-rich behavior.

The results in \cref{fig:dsrl_touch_ablations} support two conclusions. First, tactile observations improve the diffusion-steering state representation: for the same disagreement modality, solid \algdsrltacmaxinfosac curves consistently outperform their dashed \algdsrlmaxinfosac counterparts, which do not receive tactile input. Second, exploration itself is important, as the colored \algdsrltacmaxinfosac variants improve over the black SAC baselines. However, the strongest performance is obtained when tactile information is used not only as an observation, but also as part of the disagreement signal. When DSRL receives tactile observations but computes disagreement only over non-tactile modalities, performance degrades relative to tactile-disagreement variants. This indicates that touch is most useful when it directly guides epistemic exploration toward contact-rich transitions.

We include the explore-then-exploit variant as a schedule control, shown by the dotted curve in \cref{fig:dsrl_touch_ablations}. This method trains without an extrinsic reward for the first 100k environment steps, and then greedily maximizes w.r.t.~the task. Since it changes the training schedule rather than the sensing modality, we isolate it in a separate panel and do not use it for the main modality comparison.

\begin{figure}[t]
    \centering
    \includegraphics[width=0.8\linewidth]{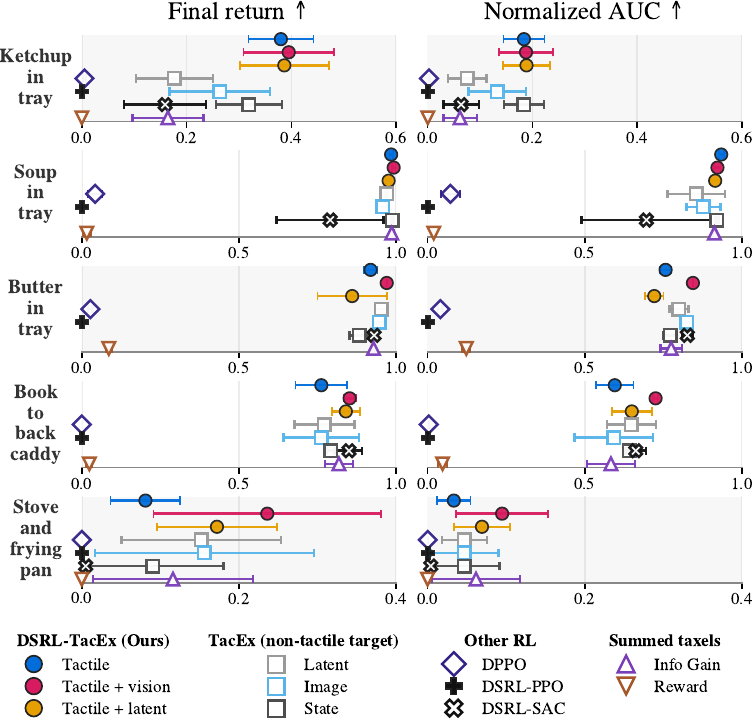}
    \caption{Additional diffusion-policy comparisons on five LIBERO tasks. Final return is the time- weighted mean success rate over the final 10\% of the logged training horizon; normalized AUC is the integral over the full horizon divided by that horizon. Points and error bars report means and standard errors across ten seeds per task. Circles and squares denote bonuses with and without explicit tactile prediction targets. Summed-taxel controls use the aggregate signal as an intrinsic reward or as a prediction target for information gain. DSRL-PPO steers a frozen policy using task reward; the DPPO adaptation updates the action-output head using tactile disagreement.}
    \label{fig:dsrl_extended_comparison}
\end{figure}

\Cref{fig:dsrl_extended_comparison} compares five LIBERO tasks: ketchup in tray (58), soup in tray (55), butter in tray (56), book to back caddy (77), and stove and frying pan (21). The added ``summed taxels" baselines either use the aggregate signal directly as a diffusion-steering reward, approximating force-magnitude intrinsic rewards, or use the ensemble disagreement when predicting the same signal as a MaxInfoRL bonus \citep{sukhija2024maxinforl}. This isolates the value of spatially resolved tactile uncertainty. DSRL-PPO and DPPO test post-training; DSRL-PPO replaces SAC with PPO while keeping the setup and interaction budget fixed. This provides the matched on-policy comparison, although DSRL-PPO does not achieve meaningful learning progress on these five tasks within the evaluated interaction budget. We additionally include DPPO~\citep{ren2024dppo}, a structurally different baseline that fine-tunes the diffusion policy itself using policy gradients.

\end{document}